\documentclass[11pt]{article}

\usepackage{natbib,amsmath,amssymb,amsfonts,dsfont,enumitem,float, bm}
\RequirePackage[colorlinks,citecolor=blue,urlcolor=blue,breaklinks]{hyperref}
\usepackage[normalem]{ulem}
\usepackage{epsfig,longtable,xcolor,algorithm,algpseudocode}

\usepackage{mathtools}

\begin{document}

\title{Considerations Across Three Cultures: \\
Parametric Regressions, Interpretable Algorithms, and Complex Algorithms}
\author{ Ani Eloyan\\
Brown University\\
  and\\
   Sherri Rose\\
   Stanford University}

\maketitle

\begin{abstract}
We consider an extension of  Leo Breiman's thesis from ``Statistical Modeling: The Two Cultures'' to include a bifurcation of algorithmic modeling, focusing on parametric regressions, interpretable algorithms, and complex (possibly explainable) algorithms.
\end{abstract}

\section{Introduction}
The relevance of the themes presented in ``Statistical Modeling: The Two Cultures'' by Leo Breiman remains twenty years later. While we could consider many categorizations of statistics culture, we posit that at least three cultures have emerged. We still have the data modeling group with regressions defined within parametric models, but the algorithmic modeling culture has, at a minimum, bifurcated with interpretable algorithms and (possibly explainable) complex algorithms \citep{rudin2019stop}. Practitioners of algorithmic modeling may develop interpretable or complex algorithms or both, depending on the needs of the scientific question. As empirically driven statisticians, we would prefer to collect data on this before putting forth a supposition, but, in lieu of such data, we also surmise that the algorithmic modeling culture has grown larger over time than Breiman's proposed 2\% of statisticians. In this commentary, we remark on several areas of increasing concern for statisticians---in all three cultures---working to solve real, substantive problems.

\section{Data}

Many statisticians are well aware that an analytic dataset is created after various forms of inclusion and exclusion criteria have been applied in collecting the data, followed by possible dropout of subjects and missingness of variables, among other issues. However, discussions regarding algorithms, such as the choice between parametric statistical approaches and more flexible machine learning techniques, often omit considerations of more elaborate forms of data preparation, and treat the provided data as observed. Ignoring such preprocessing or data engineering can lead to erroneous results that may be misleading or even harmful \citep{leek2015statistics,chen2020ethical}. This is particularly salient with recent forms of data like medical images. 

When analyzing functional magnetic resonance imaging (fMRI) data, a collection of preprocessing steps is often applied to prepare data for analyses. Some commonly used preprocessing procedures include motion reduction (as study participants often move in the MRI scanner), transformation of individual images to a common three-dimensional space for group analyses, and image smoothing \citep{ombao2016handbook}.  The choice of preprocessing steps, as well as the selection of software tools for performing preprocessing, can have a tremendous impact on the statistical analyses and resulting inference. For example, using a dataset of images collected from individuals with multiple sclerosis, \citet{eloyan2014health} showed that the results from image registration in a common template space can vary significantly depending on the choice of registration software when individuals are affected by the disease. One possible reason for these differences is that most commonly used software is developed for analyzing  data from healthy controls, hence the issues with generalizability to populations with differing brain structures. Using the outputs from these tools as fixed input data as if it were the observed data---for a parametric regression or an algorithm---without deep consideration of the implications of the preprocessing could be disastrous in a health care setting.

\section{Algorithms}

Analogous to the issues found in overlooking the nuances of data preprocessing, parametric modeling and algorithmic implementations go well beyond selecting an estimator. The reliance on software for applied problems is, again, both a necessary component in the wider adoption of important tools as well as a hindrance to good statistical practice. Various ``hidden'' assumptions and hyperparameter specifications lurk under the default settings. Here, we examine additional illustrations from imaging.

Statistical parametric mapping is widely used in fMRI studies to understand brain activation, brain functional organization, and differences in brain function \citep{friston1994statistical}. These approaches are routinely implemented in software packages used by neuroscientists and clinicians performing the analyses. However, \citet{eklund2015can} demonstrated troubling issues in applying these parametric methods with this off-the-shelf software. Specifically, that statistical parametric mapping methods are invalid for cluster-based fMRI studies. The paper started a wide discussion in the field with calls for change \citep{brown2017controversy,eklund2017reply}.

The lack of requisite detail to reproduce analyses is another major problem, particularly in algorithmic implementations. For example, deep learning \citep{goodfellow2016deep} is increasingly used in image analysis for outcome prediction, such as cancer survival or clinical outcomes in dementia. \citet{mckinney2020international} proposed a deep learning system making bold claims regarding outperforming radiologists in reading mammograms to identify cancer. Motivated by this work, \citet{haibe2020transparency} described issues with transparency and reproducibility in implementations of artificial intelligence systems for imaging-based prediction studies, e.g., missing hyperparameters from the three algorithms described by \citet{mckinney2020international}.

These examples show that even if algorithmic approaches have highly accurate performance in a specific study, it may be impossible to replicate the results or implement the same methods in other settings if sufficient details are not provided. Publicly sharing data may not be possible or appropriate due to patient privacy, however, providing similar simulated datasets paired with code can partially resolve this issue, and detailed appendices should include key hyperparameters and other algorithm attributes. 

\section{Uncertainty and Interpretation}

Quantification of uncertainty is often automatic in computational implementations of parametric statistical methods for data analysis, although it is frequently for the wrong target (e.g., standard errors for coefficients rather than prediction intervals for outcomes). Worse yet, in applications of complex or interpretable algorithms, the predicted values or assessment of accuracy is often reported without any estimates of uncertainty of the error. Regardless of which ``culture'' we fall into, as statisticians, we should be centering the role of uncertainty.

In examples discussed by \citet{breiman2001statistical}, point estimates of prediction or misclassification errors are presented and compared. Bootstrap methods \citep{efron1992bootstrap} can be implemented for estimation of prediction and confidence intervals in many nonparametric estimation approaches. However, the implementation of bootstrap methods may not be feasible for some algorithmic approaches due to computational complexity or a breakdown of the required assumptions. While many algorithm-specific methods have been developed for assessing uncertainty in these settings, such as the Bayes-by-backprop approach for neural networks \citep{blundell2015weight}, uncertainty estimates are rarely presented in applications. 

Despite the field's greater focus on algorithms for prediction, machine learning techniques have also been developed for (causal) effect estimation. Uncertainty assessment clearly has major relevance for algorithms designed to estimate effects (what Breiman refers to as ``Information''). Here, bootstrapping and influence-curve-based methods have been shown to be useful, such as in targeted maximum likelihood estimation \citep{van2011targeted,cai2019nonparametric}. Post-selection inference methods are also available for when model coefficients are the target of inference after a data-adaptive fitting process \citep[e.g.,][]{berk2013valid}. 

Appropriate interpretations of algorithmic results are crucial. Even with rigorously calculated prediction or confidence intervals, thoughtful consideration of the generalizability or transportability of the study results is essential \citep{degtiar2021review}. To what target population do we aim to make conclusions? Does this tool inform at the population level?

\section{Discussion}

The statistical literature is often behind other fields involved in data analysis, largely due to the focus on parametric model-based approaches, as discussed by \citet{breiman2001statistical}, including prioritization of ``irrelevant theory,'' and partially due to the comparatively slower speed of publication in many statistics journals (vs. computer science conference publications). In addition to expanding our focus on flexible algorithmic approaches, increasing meaningful collaborations with subject matter experts and community members is crucial. Our data are complex, particularly in the area of health (e.g., electronic health records, medical imaging, genomics). It is difficult to learn the necessary background information in isolation in order to form a plausible statistical formulation of the scientific problem and make reasonable assumptions for the analysis methods. We must also consider the potential impact of an algorithm on the communities where it may be applied \citep{vyas2020hidden,chen2020ethical,kasy2021fairness}, which also cannot be done in isolation. Statistics as a field has not embedded community-based participatory research in its overall culture to our detriment. What happens next (e.g., after this paper is published in a statistics journal) is typically not asked. Who might use this and how might they use it? Does this tool perpetuate or instigate harm? We argue that these questions, and the other considerations raised in our commentary, are critical areas of focus for all three statistics ``cultures.''

\section{Acknowledgements}
This work was supported by grant 5P20GM103645 from NIGMS and NIH New Innovator Award DP2MD012722.

  \bibliographystyle{plainnat} 
\bibliography{EloyanRose} 
\end{document}